\documentclass[twocolumn]{article}
\usepackage[utf8]{inputenc}
\usepackage{graphicx}
\usepackage[backend=bibtex]{biblatex}
\usepackage{geometry}
\usepackage{subcaption}
\usepackage{colortbl}
\usepackage[dvipsnames]{xcolor}

\definecolor{lightgray}{rgb}{0.9,0.9,0.9}

\title{Sign Language Recognition using Temporal Classification}
\author{Hardie Cate (ccate@stanford.edu)\\ Fahim Dalvi (fdalvi@cs.stanford.edu) \\ Zeshan Hussain (zeshanmh@stanford.edu)}
\date{December 11, 2015}

\begin{document}

\maketitle

\section{Introduction}
In the US alone, there are approximately 900,000 hearing impaired people whose primary mode of conversation is sign language. 
For these people, communication with non-signers is a daily struggle, and they are often disadvantaged when it comes to finding a job, accessing health care, etc.
There are a few emerging technologies aimed at overcoming these communication barriers, but most existing solutions rely on cameras to translate sign language into vocal language.
While these solutions are promising, they require the hearing impaired person to carry the technology with him/her or for a proper environment to be set up for translation.

One alternative is to move the technology onto the person’s body.
Devices like the Myo armband available in the market today enable us to collect data about the position of the user's hands and fingers over time.
Since each sign is roughly a combination of gestures across time, we can use these technologies for sign language translation.
For our project, we utilize a dataset collected by a group at the University of South Wales, which contains parameters, such as hand position, hand rotation, and finger bend, for 95 unique signs.
For each input stream representing a sign, we predict which sign class this stream falls into.
We begin by implementing baseline SVM and logistic regression models, which perform reasonably well on high-quality data.
Lower quality data requires a more sophisticated approach, so we explore different methods in temporal classification, including long short-term memory architectures and sequential pattern mining methods. 

\section{Related Work}
Several techniques have been used to tackle the problem of sign language to natural language translation. Conventionally, sign language translation involves taking an input of video sequences, extracting motion features that reflect sign language linguistic terms, and then using machine learning or pattern mining techniques on the training data. For example, Ong et al. propose a novel method called Sequential Pattern Mining (SPM) that utilizes tree structures to classify signs \cite{ong2012sign}. Their data was captured using a mobile camera system and the motion features extracted included motion of the hands, location of the sign being performed, and handshapes used. These data are collected temporally (i.e. at multiple time steps). The authors use an SPM method in order to address common issues with other models that have been used for sign language classification and translation, including unrefined feature selection and non-discriminatory learning. In general, there has been significant work done in using sequential pattern mining methods to analyze temporal data \cite{ong2012sign}\cite{papapetrou2007}. Other authors use Hidden Markov Models (HMM) on features extracted from video frames to perform the same task. Starner et. al. hypothesize that the fine movements of a signers hand are actually not required, and that coarse position and orientation of the hand are discriminative enough to classify signs \cite{starner1998real}. They also try two camera perspectives, one directly in front of the signer and the other on a cap that the person is wearing. 

Other techniques that do not involve visual input have also been studied. These techniques normally rely on data from sensors that measure features like hand positions and orientations, finger bend measures, etc. For example, Kadous uses a novel technique in his paper to improve classic machine learning algorithms by creating meta-features \cite{kadous2002temporal}. These meta-features are derived from the raw features by looking at important events in the time-series data. An example Kadous uses in his paper is the vertical maxima that a person's wrist reaches while signing. A meta-feature like this gives meaning to the "y-axis" in the data, and helps create better features. He also looks at other automatic techniques to generate these meta-features by looking for variations across time in a given dataset. Another study by Mehdi et al. proposes the use of neural networks to classify signs correctly \cite{mehdi2002sign}. They focus on signs that have distinct static shapes rather than analyzing the signs over time. Some studies have also suggested using HMMs to detect the gestures being performed. Liang et. al. employ this technique, along with a Sign Language to English language model to predict a particular sign \cite{liang1998real}. Finally, a study by Graves et al. uses strong classification to predict a sequence of labels given time-series data, rather than a single label \cite{graves2006connectionist}. They use a recurrent neural network (RNN) integrated with a softmax classifier to achieve this prediction. 

\section{Datasets}
We are primarily using two datasets from the research project by Kadous \cite{kadous2002temporal}. The first dataset is a high quality dataset, with data recorded at a frequency of 200 Hz. This dataset was recorded using two 5 dimensional flock gloves on each hand. We have access to 6-bit precision finger bend measures for each finger, as well as 14-bit precision orientation and position of each wrist in space. The second dataset is a low quality dataset, with data available for only one hand. The data was recorded at a much lower frequency of 50 Hz using the Nintendo Powerglove. We only have 2-bit precision for the finger bend measures, and data was recorded for only 4 fingers. The position of the wrist also has a lower precision of 8-bits. In addition, we have access to only one degree of rotation, which was recorded at roughly 4-bit precision. Both datasets consist of time-series data for 95 signs. In the high quality data, we have 27 instances of each sign, while in the low quality data, we have 70 instances of each sign. These instances were recorded across various sessions. The source of the high quality data was a single professional signer, while the low quality data was recorded using several signers of varying levels of proficiencies. 

\subsection{Preprocessing}
\begin{figure}
\centering
\includegraphics[width=\linewidth]{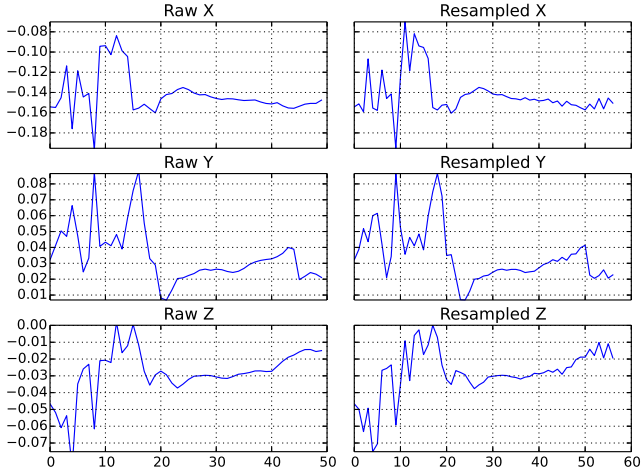}
\caption{Signal resampling}
\label{fig:resample}
\end{figure}
We perform some of the following preprocessing steps on the dataset, depending on which algorithm we are using:
\begin{enumerate}
\item \textbf{Temporal scaling:} The average number of frames for each sign is 57 frames, so we normalize all signs to this length by resampling using a fast Fourier transform. In Figure \ref{fig:resample}, the graphs on the left display readings for a single motion parameter over the lifetime of the sign, while the corresponding graphs on the right depict the resampled version. In general, we notice that there are not many differences between the original data and the resampled version. Most of the important information is retained, which supports our choice of normalization. This also helps nullify the affects of the speed at which a sign is performed, as some signers sign at a higher speed than others. Temporally scaling the data would normalize all signals to be at roughly the same speed.
\item \textbf{Spatial scaling:} Each sign in our datasets (especially in the low quality dataset) was performed at different times under different conditions. Hence, the signs were performed at varying relative positions and orientations. For example, a sign that involves moving the wrist in a parabolic motion may have its vertex at varying heights between runs. To normalize these variances, we spatially scale all the signals to be between $0$ and $1$.
\item \textbf{Time-series flattening:} For our datasets, each sample consists of a $\mathtt{NUM\_FEATURES} \times \mathtt{TIME\_STEPS}$ matrix, where each row represents one of the distinct motion parameters and each column is a particular frame. We process this data and store it in a 3-D matrix, whose dimensions are $\mathtt{NUM\_EXAMPLES} \times \mathtt{NUM\_FEATURES} \times \mathtt{TIME\_STEPS}$. For the algorithms that cannot take into account the temporal nature of the data, we transform this matrix into a flattened 2-D matrix of size $\mathtt{NUM\_EXAMPLES} \times \mathtt{(NUM\_FEATURES} \times \mathtt{TIME\_STEPS)}$ such that the first $\mathtt{NUM\_FEATURES}$ features in each row are the reading at time $t=1$, the next $\mathtt{NUM\_FEATURES}$ features correspond to time $t=2$, etc.
\end{enumerate}

\section{Technical Approach}
\subsection{Baseline}
We begin with several baseline implementations using SVM and logistic regression models. Since both of these techniques require single dimensional features, we use temporal scaling and time-series flattening. We use both linear and RBF kernels for our SVM, but the additional complexity of the RBF kernel did not improve our results significantly. For both of these models, we use one-vs-rest strategy to build classifiers for each sign. We train both of these models on 70\% of the data, and use the remaining 30\% as test instances for each of the 95 signs. We use the scikit-learn library for our SVM and logistic regression implementation \cite{scikit-learn}.

\subsection{Long-Short Term Memory}
The first complex model that we work with is a recurrent neural network, specifically the long-short term memory architecture. We choose this model because it takes into account the temporal features of our data, unlike the baseline models. We start out by trying to mimic the results of logistic regression using a simple neural network with a hidden layer that used sigmoid activation. Once we have sufficient performance, we use the same architecture for each time step, and then connect the hidden layers so that we can perform backpropagation with time. Other architectures with more layers (both fully connected and partially connected layers) were also considered. The final architecture is a three layer network. The first layer is a time-connected layer, the second layer is a fully connected dense layer for each time step, and the final layer is a dense layer that outputs a $95$-dimensional vector. We utilize mean squared error as our loss function. We use the Keras library for our LSTM implementation \cite{keras}.

\subsection{Sequential Pattern Mining}
The second technique that we try Sequential Pattern Mining. Batal et. al. have described an algorithm to perform multi-variate time series classification \cite{batal2009multivariate}. Their method is primarily a feature engineering technique that looks at the combination of the signals in an instance and outputs a binary feature vector. We can then train a standard SVM over these binary feature vectors to perform our classification. The algorithm tries to find a set of patterns in the signals that serve as a fingerprint for the class of that signal. After preprocessing (temporal and spatial scaling), SPM primarily has three steps:
\begin{enumerate}
\item \textbf{Discretization:} The first step in the algorithm is to discretize the input signals into discrete values. In our implementation, we try two different sets, {\textit{high, middle, low}} and {\textit{very high, high, middle, low, very low}}. Because of our spatial scaling, the signal values are between $0$ and $1$. We set thresholds for each of the discrete values depending on the set we are using. Thus, the discretized signal might look like the following: \texttt{HMLLHHLMMMMLH}, where $H$ is high, $M$ is middle, and $L$ is low. Subsequently, we combine all consecutive values that are equal. Hence, our example would transform into \texttt{HMLHLMLH}.
\item \textbf{Candidate Pattern generation:} The next step of the algorithm is generating patterns. A pattern is defined as a list of states, where each pair of consecutive states is connected by a relation. In our case, the states are \texttt{H:1}, which indicates that signal 1 was high. We also consider only two relations, \textit{before} and \textit{overlap}. Hence, \texttt{H:1-b;L:2} implies that the state with high value in signal 1 occurs \textit{before} another state in signal 2 that has a low value. Generating candidate patterns alternates between generating all possible patterns of a certain length $k$ and pruning them. We start with all patterns of length 1, i.e. $k=1$, then proceed to $k=2$ etc. To prune patterns, we see if the pattern appears in any of the instances of a given class. If it appears a minimum number of times (denoted by \textit{support}), we keep the pattern. Thus, we follow the approach of the Apriori algorithm for frequent item set mining which relies on the fact that any of a pattern's subpatterns appear at least as often as the pattern itself. Generating patterns of length $k+1$ involves considering all patterns of length $k$ that have the same $k-1$ states as their prefix.
\item \textbf{Binary vector creation:} Finally, having a set of patterns that remain after our candidate generation, we use a Chi-square test to rank all the patterns by their ability to distinguish between the signs. After ranking, we choose the top \texttt{MAX\_PATTERNS} patterns, where \texttt{MAX\_PATTERNS} is a hyperparameter we tune. Once we have \texttt{MAX\_PATTERNS} patterns, for each instance in our dataset, we build a binary feature vector, where a $1$ in position $i$ indicates that pattern $i$ occurs in that instance. 
\end{enumerate}

\section{Results and Analysis}
\subsection{Experiments}
With each of our models, we try several experiments. Specifically, for our baseline SVM model, we try various kernels. We also try regularizing both our baseline SVM and logistic regression models to prevent overfitting on the data. For the LSTM architecture, we experiment with several activation functions for each layer and try adding intermediate layers like Dropout to prevent overfitting. Finally, for our SPM approach, we have several hyperparameters to tune such as the window size, length of patterns generated during candidate generation and the minimum support required for each candidate pattern. We also tune parameters such as the regularization and feature vector length on which we train the SVM. Finally, since SPM involves discretizing the signal, we try two different implementations: discretizing the raw signal itself and discretizing the rate of change in the signal.

\subsection{Results}
\begin{table}[b]
    \centering
    \caption{Algorithm performance on the low quality dataset}
    % \resizebox{\columnwidth}{!} {
        \begin{tabular}{|c|c|c|c|c|}
        \hline
         & SVM & Log. Reg. & LSTM & SPM\\\hline
        Precision & $0.566$ & $0.444$ & $0.109$ & $0.075$  \\
        Recall & $0.550$ & $0.444$ & $0.091$ & $0.076$ \\
        F1 & $0.549$ & $0.436$ & $0.066$ & $0.065$ \\ 
        Training error & $0.001$ & $0.179$ & $0.852$ & $0.526$ \\
        Testing error & $0.450$ & $0.556$ & $0.908$ & $0.895$ \\ \hline
        \end{tabular}
    % }
    \label{table:comparison}
\end{table}
The baseline models perform very well on the high quality data. Both the SVM and the logistic regression models give us a test error of $5.8\%$ on the high quality dataset. Since we already have good performance on the high quality dataset, we focus on the low quality dataset for the remainder of the project. The results on the low quality dataset are shown in table \ref{table:comparison}. 

%The logistic regression model performs slightly worse, with test errors of $5.8\%$ and $55.6\%$ for the two datasets. We note that both models perform significantly better on the high quality dataset than the low quality dataset, which is expected since the low quality dataset has significantly less features and the data itself is noisier. We performed several diagnostics to explore the high accuracy of the models on the high quality dataset, including calculating confusion matrices, running ablation tests, and mapping the feature space. Below, we report our analysis for the SVM model on the high quality data set. 
\begin{figure}[t]
\centering
\begin{subfigure}[b]{0.48\linewidth}
        \includegraphics[trim=100 22 80 39, clip,width=1.0\linewidth]{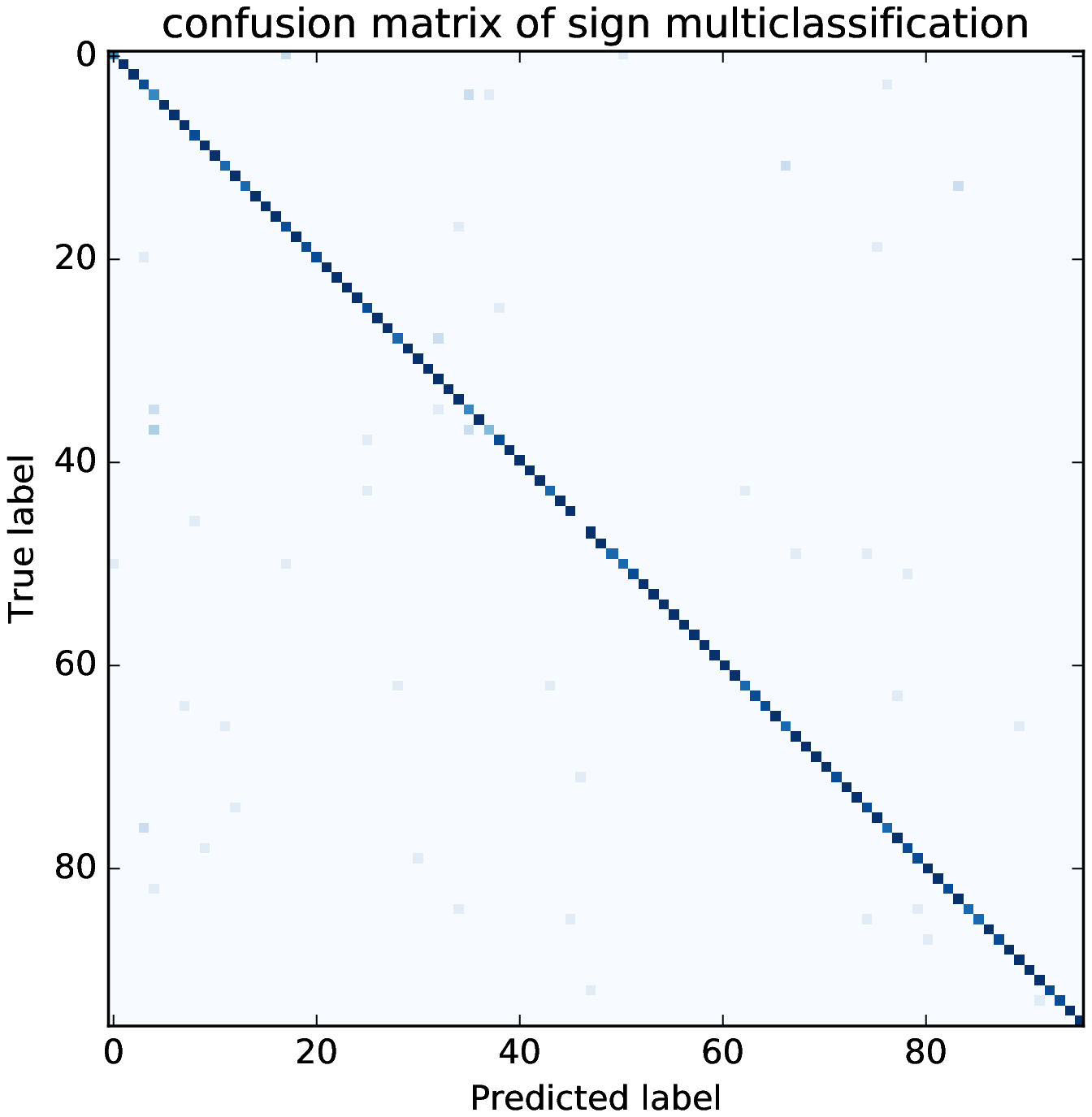}
        \caption{High Quality}
\end{subfigure}%
~
\begin{subfigure}[b]{0.48\linewidth}
        \includegraphics[trim=100 22 80 39, clip,width=1.0\linewidth]{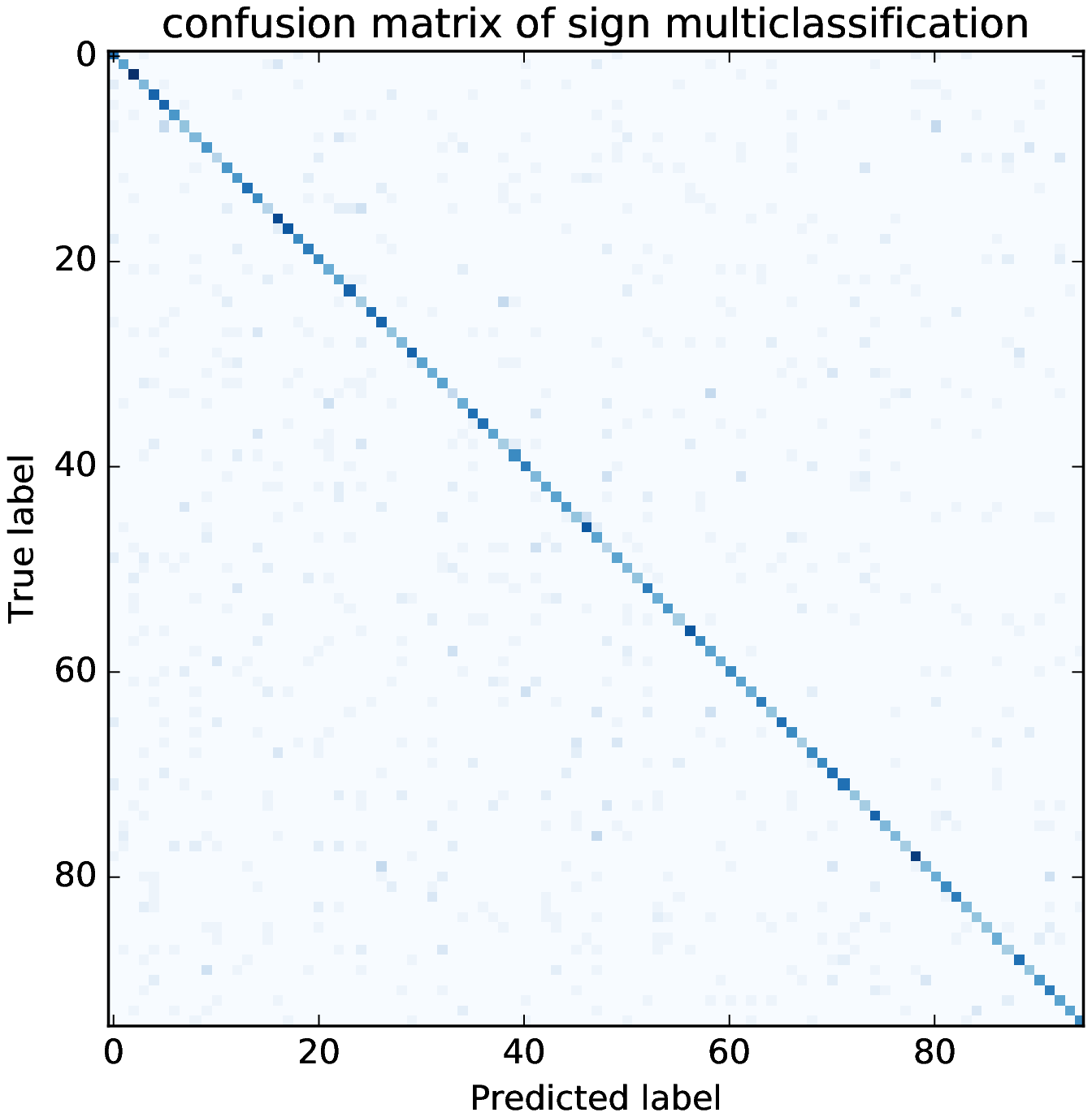}
        \caption{Low quality}
\end{subfigure}%
\vspace{-2mm}
\caption{Confusion matrix for SVM}
\label{fig:confusion_matrix_svm}
\end{figure}

First, we note that both our training and testing errors using SVM are very low, so our model is not suffering from overfitting (see Table \ref{table:comparison}). Additionally, all other metrics, including precision, recall, and F1 score are very high, suggesting a high, but also precise, level of performance. This theory is substantiated by the confusion matrix for the SVM (see Fig. \ref{fig:confusion_matrix_svm}a), which shows that the classifier is not confusing a sign with some other sign. The clear blue line down the diagonal is evidence of this claim. Note that the confusion matrix as well as the metrics on the low quality dataset are much worse than those on the high quality dataset. In general, our classifier confuses similar signs more often, which is expected because there might not have been enough features to distinguish these signs.

Unlike the baseline models, the LSTM results are poor on the low quality dataset. Although we initially expect the LSTM to perform better, as it takes into account the temporal nature of the data, the performance does not change much even after varying architectures for the LSTM. We hypothesize that this is because of some key assumptions that the standard LSTM model makes that do not apply to our data. In the standard LSTM model, backpropagation through time is done at every time step. This is usually acceptable for time series data since we normally want to predict the value at the next consecutive time step. However, in our case, backpropagating at each step leads to a poorer model, since we are not trying to predict the next value in the signal (eg. next hand position, orientation etc.), but rather we want to classify the entire signal as one unit. Hence, we would like to only backpropagate at the final timestep to achieve a better model. Since building a custom LSTM model would be time consuming, we pursue SPM as it is more promising for our particular task.

The SPM model performs slightly better than the LSTM model. However, with around $10\%$ accuracy, the model is not very strong. For the best result, we had a window size of 2, minimum support of 20 and a maximum pattern length of 2. We posit that because of such limiting values, the patterns we generate are not very long. Hence, we are losing a lot of distinguishing information by not having longer patterns. Although slightly longer patterns (window size $\sim$ 10, maximum pattern length $\sim$ 15) give us poorer performance, we think that using much longer patterns (window size $\sim$ 20, maximum pattern length $\sim$ 30) would give us a better result. Unfortunately, our implementation runtime increases exponentially as we increase these parameters, and hence we were limited on the maximum length up to which we could increase our generated patterns. Another hypothesis we have for the generally poor performance is that the SPM algorithm gives us a set of patterns it thinks are most distinguishing. After this, we check if each of the patterns occur in the instances to build our feature vector, but we do not take into account where or how many times each of these patterns occur in the instances, thus losing some more information in this process.

\subsection{Analysis}
\begin{figure}[h]
    \centering
    \includegraphics[trim=100 40 80 80, clip, width=0.8\linewidth]{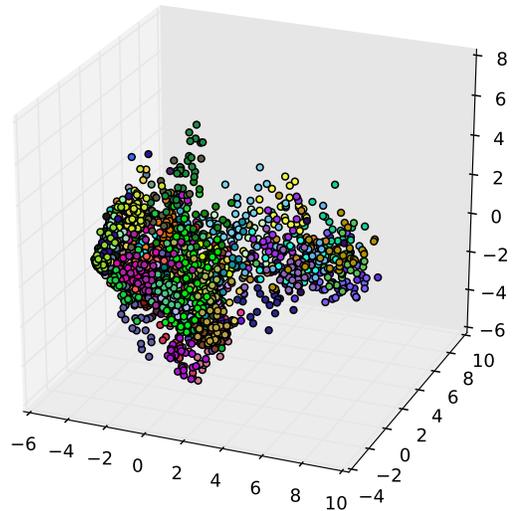}
    \caption{Feature space reduced to three dimensions}
    \label{fig:feature_space}
\end{figure}

To explain the variances in out results between the different models, we decide to analyze the feature space, importance of each feature and the effect of various hyperparameters on our models. 

Given that the baseline models perform better than the other models, we decide to analyze the flattened feature space and see if it was truly separable in the high dimensional space.  We hypothesize that each class has its own cluster in the high dimensional space and is far from other classes. One way to confirm this is to use PCA to reduce the feature space to 3 dimensions and plot the data. We see in Figure \ref{fig:feature_space} that examples from the same class (indicated by similar colors) cluster together, which indicates the presence of the clusters in the higher dimensional space. 

Next, we want to see which of the features are most discriminative, so that we could restrict the features we were using the the more complex models to save on runtime. We run two rounds of ablation tests on the SVM to determine which features were the most significant contributors to the overall performance.  The first round removes each feature independently and measures the results on the data without that one feature, while the second round removes an increasing number of features. All of these tests are run on the low quality dataset. From the results of these tests, we see that removing the position features of the hand results in a significant increase in test error (see Table \ref{table:ablation}). Additionally, the largest jump in error for the second set of ablation tests occurs when we remove the position and rotation features. Removing the finger features does not have a significant impact on the performance of the model, suggesting that the position and rotation features are the most distinguishing features between the signs.

\begin{table}[h]
    \centering
    \caption{Ablation test results}
    \label{table:ablation}
    % \resizebox{\columnwidth}{!} {
        \begin{tabular}{|l|c|}
        \hline
        Removed features & Test error \\ \hline
        None & 0.450 \\ \arrayrulecolor{gray}\hline\arrayrulecolor{black}
        \rowcolor{lightgray} POS & 0.786 \\ \arrayrulecolor{gray}\hline\arrayrulecolor{black}
        ROT & 0.519 \\ \arrayrulecolor{gray}\hline\arrayrulecolor{black}
        F1 & 0.491 \\ \arrayrulecolor{gray}\hline\arrayrulecolor{black}
        F2 & 0.472 \\ \arrayrulecolor{gray}\hline\arrayrulecolor{black}
        F3 & 0.489 \\ \arrayrulecolor{gray}\hline\arrayrulecolor{black}
        F4 & 0.458 \\ \arrayrulecolor{gray}\hline\arrayrulecolor{black}
        POS, ROT & 0.881 \\ \arrayrulecolor{gray}\hline\arrayrulecolor{black}
        POS, ROT, F1 & 0.896 \\ \arrayrulecolor{gray}\hline\arrayrulecolor{black}
        POS, ROT, F1, F2 & 0.898 \\ \arrayrulecolor{gray}\hline\arrayrulecolor{black}
        POS, ROT, F1, F2, F3 & 0.929 \\ \hline
        \end{tabular}
    % }
    \caption*{\small{Here, \texttt{POS} and \texttt{ROT} refer to the position and orientation of the right wrist respectively. \texttt{F\#} refers to the fingers on the hand. The ordering of the fingers is thumb, index, middle, ring.}}
\end{table}

\begin{figure}[h]
    \centering
    \includegraphics[trim=50 0 30 0, clip,width=0.95\linewidth]{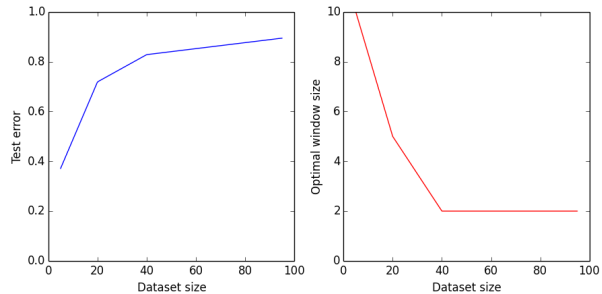}
    \caption{Performance of SPM with varying dataset sizes}
    \label{fig:spm_performance}
\end{figure}

% TODO Correct
Furthermore, we perform extensive hyperparameter tuning for our SPM model. Since our hyperparameters space is quite large, we use a procedure akin to coordinate ascent to find the optimal set of hyperparameters. For the larger models, we also start with a small number of signs to build an intuition on the affect of varying each hyperparameter, and then slowly expand our training/test datasets to include more signs. As shown in Figure \ref{fig:spm_performance}, the test error increases as we include more signs in our analysis. We started with 5 signs and progressively added sets of signs, and for each set we computed the hyperparameters that gave us the best accuracy. As we can see, the window size reduces as we increase the number of signs. One reason for this may be as we increase the number of signs, the probability of us seeing a pattern again increases if the window size is held constant. Since we want to choose the patterns that are most discriminative, a lower window size leads to better performance with a high number of examples. We also notice that the optimal window size and maximum pattern lengths are quite small ($<$ 10) for all sets of signs we tried, indicating that the instantaneous motions in each sign serve as discriminative features, rather than longer patterns.

As our final analysis, we also try concatenating the feature vectors we get from the SPM algorithm along with the raw flattened features. Using a small subset of these features ($\sim$50), we found that we get a $1\%$ bump in accuracy over our baseline SVM. 

\section{Conclusions and Future Work}
In this paper, we study and apply machine learning techniques for temporal classification, specifically the multi-variate case. Although the results obtained are not very high, we believe that a more efficient implementation of the algorithms can yield bigger and more complex models that will perform well. 

In the future, we plan to improve the implementation behind SPM to build better models. We may also consider implementing a custom LSTM model that removes the assumptions of the technique that do not apply to our data. Finally, we would also like to use a device available in the market today, namely the Myo armband, to record our data, and try our models on this data. Even though the data that will be collected will not be exactly the same as our current data, we believe that the techniques we have tried and implemented are general enough for them to work well on the new data. 

Most importantly, we have shown that at least with high quality data, it is indeed possible for us to translate sign language into text. We hope that some day this will enable hearing impaired people to communicate more effortlessly with the rest of the society.
\newpage
\section{References}
\printbibliography[heading=none]
\end{document}